\documentclass[journal,onecolumn]{IEEEtran}
\IEEEoverridecommandlockouts

\usepackage{cite}
\usepackage{titlesec}
\usepackage{amsmath,amssymb,amsfonts}
\usepackage{algorithmic}
\usepackage{graphicx}
\usepackage{textcomp}
\usepackage{xcolor}
\usepackage{hyperref}
\usepackage{enumitem}
\usepackage{subfig}
\usepackage[table]{xcolor}
\usepackage{pifont}
\usepackage{float}
\usepackage[table]{xcolor}
\usepackage{colortbl}
\usepackage{graphicx}
\usepackage{float}
\definecolor{low}{rgb}{1,1,1}
\definecolor{high}{rgb}{0.3,0.6,1}
\usepackage{pgf}

\newif\ifpreprint
\preprinttrue  

\begin{document}

\title{Hy-Facial: Hybrid Feature Extraction by Dimensionality Reduction Methods for Enhanced Facial Expression Classification}

\author{
\begin{minipage}[t]{0.3\textwidth}
\centering
1\textsuperscript{st} Xinjin Li\textsuperscript{*,\dag}\\
\textit{Department of Computer Science} \\
\textit{Columbia University} \\
New York, NY, USA \\
li.xinjin@columbia.edu
\end{minipage}
\hfill
\begin{minipage}[t]{0.3\textwidth}
\centering
1\textsuperscript{nd} Yu Ma\textsuperscript{*}\\
\textit{School of Computer Science} \\
\textit{Carnegie Mellon University (CMU)} \\
Pittsburgh, PA, USA \\
yuma13926@gmail.com
\end{minipage}
\hfill
\begin{minipage}[t]{0.3\textwidth}
\centering
3\textsuperscript{rd} Kaisen Ye\\
\textit{Chu Kochen Honors College} \\
\textit{Zhejiang University} \\
Hangzhou, Zhejiang, China \\
yekaisen00@gmail.com
\end{minipage}

\vspace{1cm}

\begin{minipage}[t]{0.3\textwidth}
\centering
4\textsuperscript{th} Jinghan Cao\\
\textit{Department of Computer Science} \\
\textit{San Francisco State University} \\
Seattle, WA, USA \\
jcao3@alumni.sfsu.edu
\end{minipage}
\hfill
\begin{minipage}[t]{0.3\textwidth}
\centering
5\textsuperscript{th} Minghao Zhou\\
\textit{Department of Computer Science} \\
\textit{Columbia University} \\
New York, NY, USA \\
mz2962@columbia.edu
\end{minipage}
\hfill
\begin{minipage}[t]{0.3\textwidth}
\centering
6\textsuperscript{th} Yeyang Zhou\\
\textit{Department of Computer Science} \\
\textit{UC San Diego (UCSD)} \\
La Jolla, CA, USA \\
yeyang-zhou@ucsd.edu
\end{minipage}

\vspace{0.5cm}

\textsuperscript{*}Equal contribution \quad 
\textsuperscript{\dag}Corresponding author
}

\maketitle

\begin{abstract}

Facial expression classification remains a challenging task due to the high dimensionality and inherent complexity of facial image data. This paper presents Hy-Facial, a hybrid feature extraction framework that integrates both deep learning and traditional image processing techniques, complemented by a systematic investigation of dimensionality reduction strategies. The proposed method fuses deep features extracted from the Visual Geometry Group 19-layer network (VGG19) with handcrafted local descriptors and the scale-invariant feature transform (SIFT) and Oriented FAST and Rotated BRIEF (ORB) algorithms, to obtain rich and diverse image representations. To mitigate feature redundancy and reduce computational complexity, we conduct a comprehensive evaluation of dimensionality reduction techniques and feature extraction. Among these, UMAP is identified as the most effective, preserving both local and global structures of the high-dimensional feature space. The Hy-Facial pipeline integrated VGG19, SIFT, and ORB for feature extraction, followed by K-means clustering and UMAP for dimensionality reduction, resulting in a classification accuracy of 83. 3\% in the facial expression recognition (FER) dataset. These findings underscore the pivotal role of dimensionality reduction not only as a pre-processing step but as an essential component in improving feature quality and overall classification performance.
\end{abstract}


\section{Introduction}

\ifpreprint

\fi

Facial expression recognition (FER) plays a vital role in human-computer interaction, mental health monitoring, and behavioral analysis. Yet, it remains a challenging task due to the high dimensionality of facial image data and the subtle nature of expression-related variations. Traditional approaches often struggle to balance computational efficiency and classification performance, particularly when working with complex and high-dimensional visual features.

Recent advancements in deep learning have enabled more powerful feature representations for image classification. Models such as VGG19 \cite{simonyan2015very} have shown strong capabilities in capturing high-level global semantics. Simultaneously, handcrafted feature descriptors like SIFT \cite{lowe2004distinctive} and ORB \cite{rublee2011orb} continue to provide valuable local geometric information that complements deep features. However, despite these advances, effectively reducing the dimensionality of such fused features remains an open problem. A naive application of dimensionality reduction techniques can lead to loss of structural information, degraded model performance, or increased sensitivity to noise.

Prior research can be broadly categorized into two directions: (i) combining handcrafted features with linear reduction methods like PCA \cite{Jolliffe2002}, and (ii) utilizing deep features with nonlinear methods such as t-SNE \cite{Maaten2008} and UMAP \cite{McInnes2018}. While each offers advantages, existing literature rarely explores a holistic integration of global and local features followed by systematic experimentation across a wide range of dimensionality reduction strategies and configurations.

In this work, we propose \textbf{Hy-Facial}, a hybrid feature extraction and dimensionality reduction framework tailored for facial expression recognition. Our approach incorporates both global features extracted from the pretrained VGG19 network and local descriptors derived from SIFT and ORB. To compress the resulting high-dimensional fused features, we employ K-means clustering for class-wise pattern selection, followed by various dimensionality reduction techniques such as PCA, t-SNE, UMAP, MDS, IsoMap, and LLE. A critical emphasis is placed on understanding how each reduction method interacts with the data structure, classifier type, and target dimensionality.

Our primary contributions are 3-fold:

\begin{itemize}
    \item \textbf{Hybrid Feature Representation:} We design and evaluate a multi-source feature fusion pipeline that jointly captures global semantics and local structural cues. By integrating VGG19 with SIFT and ORB, our method enables the model to recognize both coarse and fine-grained aspects of facial expressions.
    \item \textbf{Comprehensive Dimensionality Reduction Study:} We conduct a rigorous empirical study across multiple reduction methods, dimensional settings, and classifiers (RF, KNN, MLP). This includes evaluating how different techniques preserve data structure and influence classification performance. We further analyze each method’s sensitivity to noise, local manifold fidelity, and suitability for facial image characteristics, yielding insights that extend beyond this specific task.
    \item \textbf{Hy-Facial Pipeline} Extensive experiments on the FER-Plus dataset demonstrate that our approach achieves a classification accuracy of \textbf{83.3\%} using Random Forest with UMAP, outperforming baseline and conventional feature configurations. The findings confirm that carefully designed dimensionality reduction pipelines—when combined with complementary feature representations—play a decisive role in advancing FER performance. Our methodology provides a generalizable blueprint for integrating dimensionality reduction into high-dimensional image classification tasks.
\end{itemize}

\section{Related Works}

\ifpreprint
Recent studies in medical imaging and multimodal analysis have shown the effectiveness of advanced feature extraction, ensemble learning, and prompt-based strategies for tasks such as MRI reconstruction, disease prediction, biomolecular modeling, and clinical trial analysis. These works underline the value of cross-domain techniques for improving interpretability and robustness in health-related classification \cite{bian2024diffusion,yu2025crisp,yu2025prnet,yu2024scnet,wang2025applications,wang2025systematic,zhong2025comparative,lu2024uncertainty,wang2024twin,niu2025decoding,wang2025financial,Li_Wang_Chen_2024,DBLP:journals/tnn/ZhangHLDCW25,DBLP:conf/aaai/Zhang0LXCCW25}.  
In computer vision, feature extraction and object detection remain central to classification. Prior research has addressed semantic-guided editing, texture and point cloud representations, blur-robust descriptors, and lightweight traffic sign detection, offering insights for more discriminative facial expression features \cite{chen2022harnessing,liu2024difflow3d,liu2023regformer,zhou2023fastpillars,zhou2024information,zhao2024balf,bellavia2024image,edstedt2024dedode,zhang2024yoloppa,10679029}.  
Another line of work emphasizes efficient learning frameworks, including pruning, knowledge distillation, LoRA-based fine-tuning, and reinforcement learning. These approaches stress scalability and efficiency, which are critical for deploying deep models in large-scale facial expression analysis \cite{li2025miv,li2025catp,Zhang2025TimeLLaMA,Zhang2025SensitivityLoRA,Wang2025OneImage,liuutility,liu2023meta,liu2025gatedmultimodalgraphlearning,lu2025machinelearningsyntheticdata}.  
Finally, advances in autonomous driving and spatio-temporal modeling have introduced scanpath prediction, 360° image enhancement, and attention-guided representation learning. These methods illustrate how temporal dynamics and attention mechanisms enhance recognition in complex environments, offering transferable insights for dynamic facial expression analysis \cite{zeng2025FSDrive,wang2024scantd,wang2025target,10768359,WANG2024100522,li_2024_knowledge}.
\fi

Recent advances in image classification and facial expression recognition (FER) have underscored the importance of combining feature extraction techniques with dimensionality reduction strategies. Based on the nature of feature representation and reduction schemes, prior works can generally be categorized into two primary groups: (1) linear reduction with handcrafted descriptors, and (2) nonlinear reduction with deep representations.

\subsection{Linear Reduction With Handcrafted Descriptors}

A significant portion of early FER systems relied on handcrafted descriptors such as Scale-Invariant Feature Transform (SIFT)~\cite{lowe2004distinctive} and Oriented FAST and Rotated BRIEF (ORB)~\cite{rublee2011orb}, which are known for their robustness to illumination and scale changes. These descriptors capture key local patterns and are often used in conjunction with Principal Component Analysis (PCA)\cite{Jolliffe2002, Tharwat2016} to reduce the dimensionality of feature vectors prior to classification.

PCA could transform a set of correlated variables into a smaller set of uncorrelated variables called principal components. It does this by finding the directions (principal components) that maximize the variance in the data. Mathematically, PCA is performed by computing the eigenvectors and eigenvalues of the covariance matrix of the data. The objective is to maximize the variance captured by each principal component, which can be expressed as:

\begin{equation}
	\max \left( \mathbf{w}^T \mathbf{S} \mathbf{w} \right)
\end{equation}
subject to \( \| \mathbf{w} \| = 1 \), where \( \mathbf{S} \) is the covariance matrix of the data, and \( \mathbf{w} \) is the eigenvector corresponding to the principal component.

The principal components are the eigenvectors of the covariance matrix, and the eigenvalues represent the amount of variance captured by each component. The data is projected onto the first few principal components to reduce its dimensionality while preserving as much variance as possible.

Although this combination has demonstrated reasonable performance on constrained datasets, its main limitations lie in its inability to model complex nonlinear structures in facial data. PCA, as a linear projection method, primarily captures variance in the data but fails to preserve discriminative structure when features lie on nonlinear manifolds. Moreover, handcrafted features, while interpretable, are often less effective than learned representations when applied to unconstrained environments with high intra-class variability.

\subsection{Nonlinear Reduction With Deep Representations}

With the emergence of deep convolutional neural networks (CNNs), particularly architectures such as VGG19 \cite{simonyan2015very}, feature extraction has transitioned towards learned representations that offer superior performance on large-scale FER tasks. These high-dimensional feature embeddings are typically passed through nonlinear dimensionality reduction methods such as t-distributed Stochastic Neighbor Embedding (t-SNE)\cite{Maaten2008} and Uniform Manifold Approximation and Projection (UMAP)\cite{McInnes2018} to produce more compact, discriminative representations.

\subsubsection{t-SNE}

t-SNE (t-distributed Stochastic Neighbor Embedding) \cite{Maaten2008} is designed for visualizing high-dimensional data. It models pairwise similarities in both high- and low-dimensional spaces and minimizes their divergence. The objective is to minimize the Kullback-Leibler (KL) divergence:

\begin{equation}
\text{KL}(P | Q) = \sum_{i} \sum_{j} p_{ij} \log \frac{p_{ij}}{q_{ij}}
\end{equation}
where $p_{ij}$ and $q_{ij}$ represent similarities in the high- and low-dimensional spaces, respectively.

t-SNE maps data to a low-dimensional space (typically 2D or 3D), preserving local structure and revealing clusters and complex patterns.

\subsubsection{UMAP}

UMAP (Uniform Manifold Approximation and Projection) \cite{McInnes2018} aims to preserve both local and global structure. It constructs a high-dimensional graph of nearest neighbors, then optimizes a low-dimensional representation. The objective minimizes cross-entropy:

\begin{equation}
C = \sum_{i,j} \left( p_{ij} \log \frac{p_{ij}}{q_{ij}} + (1 - p_{ij}) \log \frac{1 - p_{ij}}{1 - q_{ij}} \right)
\end{equation}
where $p_{ij}$ and $q_{ij}$ represent neighbor probabilities in high- and low-dimensional spaces.

UMAP is computationally efficient and produces meaningful representations that capture both local and global structures.

\subsubsection{Isometric Mapping (IsoMap)}

IsoMap \cite{Tenenbaum2000} extends classical MDS by preserving geodesic distances between points on a nonlinear manifold. It computes the shortest paths between all point pairs and performs MDS at these distances. The geodesic distance is approximated by:

\begin{equation}
d_{ij} = \min \left( \sum_{k=1}^{n-1} d(x_k, x_{k+1}) \right)
\end{equation}

IsoMap embeds data into a lower-dimensional space, maintaining the global manifold structure, and is effective for data on smooth low-dimensional manifolds.

\subsubsection{Multidimensional Scaling (MDS)}

MDS \cite{Kruskal1964} represents high-dimensional data in lower dimensions by preserving pairwise distances. The goal is to minimize the stress function:

\begin{equation}
\text{Stress} = \sqrt{\sum_{i<j} (d_{ij} - \hat{d}_{ij})^2}
\end{equation}
where $d_{ij}$ is the original distance, and $\hat{d}_{ij}$ is the low-dimensional counterpart.

MDS adjusts point positions to minimize stress, preserve relative distances, and aid visualization.

\subsubsection{Locally Linear Embedding (LLE)}

LLE \cite{Roweis2000} preserves local structure by assuming each point and its neighbors lie on a locally linear patch. Each point is reconstructed as a weighted sum of its neighbors by minimizing:

\begin{equation}
\min \sum_{i} | \mathbf{x}_i - \sum_j W_{ij} \mathbf{x}_j |^2
\end{equation}
subject to $\sum_j W_{ij} = 1$.

After computing weights $W_{ij}$, LLE finds a lower-dimensional embedding that preserves these relationships, maintaining the local geometry.

\subsection{Motivation}

Therefore, for classical approaches that leverage handcrafted descriptors combined with linear reduction techniques like PCA, such methods often fail to capture complex nonlinear variations inherent in facial expression data. Deep learning-based features, which have been integrated with nonlinear dimensionality reduction methods such as t-SNE and UMAP, preserve local structures more effectively. They suffer from limited global continuity and interpretability.

Given these observations, our study, Hy-Facial, is a hybrid method that fuses deep and handcrafted features, followed by clustering and nonlinear reduction, to enhance facial expression classification. The integrated approach achieves improved accuracy by preserving both global semantics and local structural details in a compact feature space. 

\section{Methodology}

In this research, we propose a comprehensive image classification pipeline that integrates feature extraction, feature selection, dimensionality reduction, and classification. The workflow begins with multi-source feature extraction using a deep learning model combined with traditional computer vision algorithms, allowing the system to capture both global and local image characteristics. Subsequently, clustering is employed as a feature selection mechanism to identify and emphasize the most representative patterns in the extracted features. After this, a critical step in the pipeline is dimensionality reduction, which involves compressing high-dimensional features into a lower-dimensional space. This step is essential not only for reducing computational complexity and mitigating the curse of dimensionality but also for enhancing the classifier’s ability to discriminate between subtle differences in image data. Finally, the compact representations are fed into a classifier for prediction. 
\subsection{Overview}
\begin{figure}[htbp]
	\centering
	\includegraphics[scale=0.4]{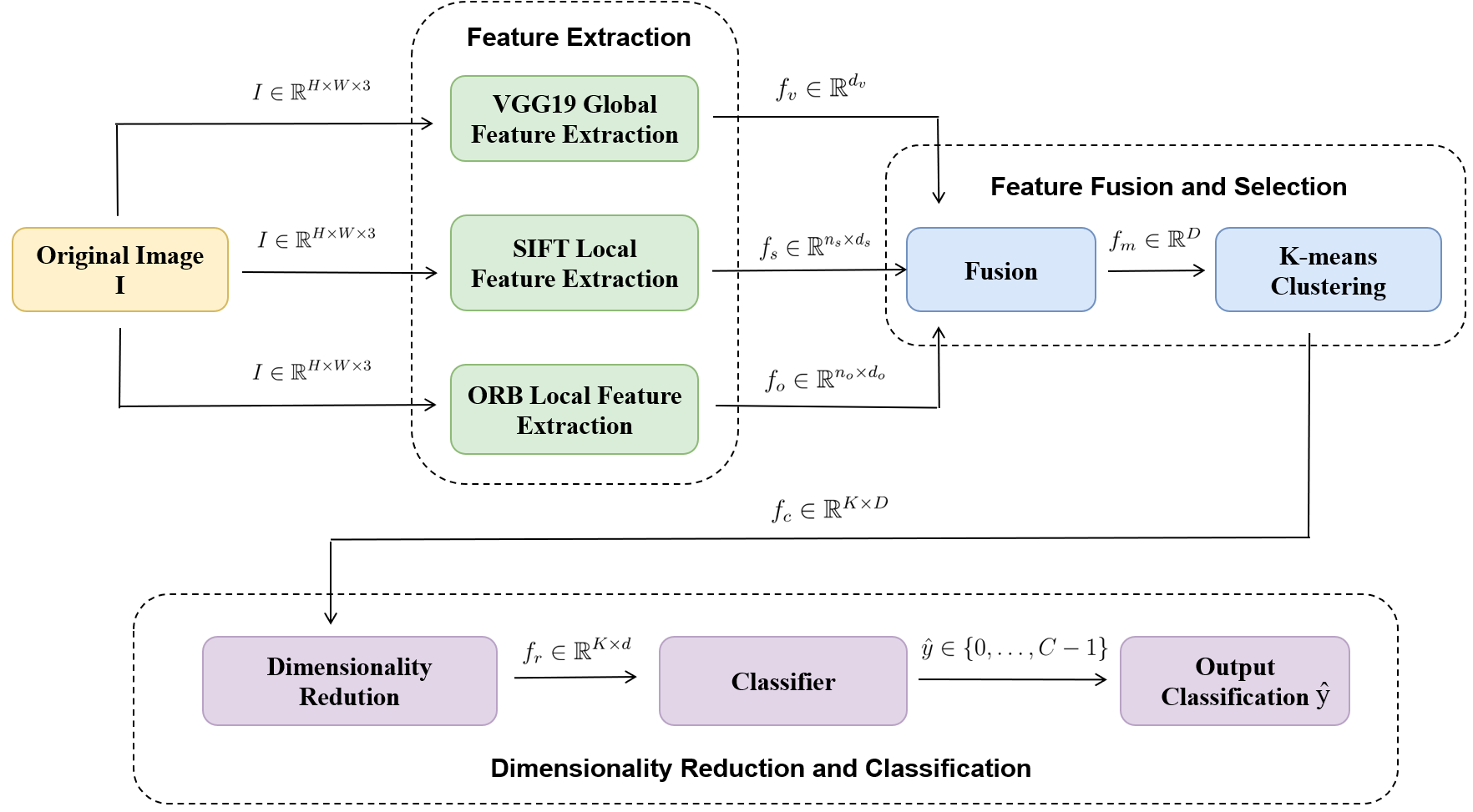}
	\caption{Pipeline of the proposed model}
	\label{fig:rf_classifier}
\end{figure}

This work addresses a multi-class image classification problem, where each input is a raw RGB image that needs to be assigned a label from a predefined category set. Formally, given an image $I$ defined over the pixel space:

\begin{equation}
I \in \mathbb{R}^{H \times W \times 3},
\end{equation}

Where $H$ and $W$ denote the height and width of the image, our goal is to learn a mapping:

\begin{equation}
f: \mathbb{R}^{H \times W \times 3} \rightarrow \{0, 1, \dots, C-1\},
\end{equation}

that assigns $I$ a discrete label $\hat{y} = f(I)$ among $C$ possible classes.

To accomplish this, we propose a four-stage framework that progressively transforms the raw image into a compact and discriminative feature representation suitable for classification. The process can be summarized as follows:

\begin{align}
I &\xrightarrow{\text{Feature Extraction}} \mathbf{f}_{\mathrm{m}} \in \mathbb{R}^{D} \\
  &\xrightarrow{\text{Feature Selection (K-means)}} \mathbf{f}_{\mathrm{c}} \in \mathbb{R}^{K \times D} \\
  &\xrightarrow{\text{Dimensionality Reduction}} \mathbf{f}_{\mathrm{r}} \in \mathbb{R}^{K \times d} \\
  &\xrightarrow{\text{Classifier (RF, KNN, or MLP)}} \hat{y} \in \{0, 1, \dots, C-1\}
\end{align}

This formalizes the entire method as a deterministic pipeline:
\[
\hat{y} = f_{\text{classifier}} \circ f_{\text{DR}} \circ f_{\text{select}} \circ f_{\text{extract}}(I)
\]

Where each transformation module is described in detail in the preceding subsections. This ensures that the input image is systematically transformed into a semantically meaningful, low-dimensional representation suitable for robust classification.

\subsection{Multi-source Feature Extraction}
Feature extraction is a critical initial step aimed at capturing comprehensive and discriminative image characteristics. To achieve this, we combine deep learning-based features with handcrafted descriptors to exploit complementary strengths.

\subsubsection{VGG19}

VGG19 \cite{simonyan2015very} is a deep convolutional neural network composed of 19 layers, known for using small \(3 \times 3\) convolutional kernels to progressively extract hierarchical features from input data. Given an input image \( I \in \mathbb{R}^{H \times W \times C} \), the core layer-wise operation is defined as:

\begin{equation}
Z^l = W^l A^{l-1} + b^l, \quad A^l = \sigma(Z^l)
\end{equation}

Where \( A^0 = I \), \( W^l \) and \( b^l \) are the weights and biases of the \( l \)-th layer, and \( \sigma(\cdot) \) is a non-linear activation function (e.g., ReLU). Through successive layers of convolution and pooling, the network transforms the input into a compact global feature vector:

\begin{equation}
\mathbf{f}_{\mathrm{v}} := A^L \in \mathbb{R}^{d_{\mathrm{v}}}
\end{equation}

Where \( L \) denotes the final layer used for feature extraction. In our work, we utilize the pretrained VGG19 up to this layer, treating \( \mathbf{f}_{\mathrm{v}} \) as the global representation of the input image for downstream tasks.

\subsubsection{SIFT}

Scale-Invariant Feature Transform (SIFT) \cite{lowe2004distinctive} is a classical algorithm for detecting and describing local features in images. It identifies keypoints by locating extrema in the Difference-of-Gaussian (DoG) scale space:

\begin{equation}
D(x, y, \sigma) = \left[ G(x, y, k\sigma) - G(x, y, \sigma) \right] * I(x, y)
\end{equation}

where \( G(x, y, \sigma) \) is a Gaussian filter at scale \( \sigma \), \( k \) is a constant factor, and \( I(x, y) \) is the input image. For each keypoint, a local descriptor is computed by aggregating gradient orientations in its neighborhood. The final set of local descriptors is represented as:

\begin{equation}
\mathbf{f}_{\mathrm{s}} \in \mathbb{R}^{n_{\mathrm{s}} \times d_{\mathrm{s}}}
\end{equation}

where \( n_{\mathrm{s}} \) is the number of detected keypoints, and \( d_{\mathrm{s}} \) is the dimensionality of each descriptor.

\subsubsection{ORB}

Oriented FAST and Rotated BRIEF (ORB) \cite{rublee2011orb} is an efficient binary descriptor for local feature extraction. It first detects keypoints using the FAST detector and then computes orientation-invariant binary descriptors based on the BRIEF operator. Let \( I(x, y) \) be the input image; for each detected keypoint, a binary descriptor is constructed by comparing the intensity values of pixel pairs within a local patch. The final local feature representation is:

\begin{equation}
\mathbf{f}_{\mathrm{o}} \in \mathbb{R}^{n_{\mathrm{o}} \times d_{\mathrm{o}}}
\end{equation}

where \( n_{\mathrm{o}} \) is the number of keypoints and \( d_{\mathrm{o}} \) is the length of each binary descriptor.

\subsubsection{Multi-source Feature Fusion}
The fusion of global features with multiple local descriptors aims to create a comprehensive and discriminative feature set. The unified feature representation is achieved through concatenation:

\begin{equation}
\mathbf{f}_{\mathrm{m}} = [\mathbf{f}_{\mathrm{v}}, \text{Flatten}(\mathbf{f}_{\mathrm{s}}), \text{Flatten}(\mathbf{f}_{\mathrm{o}})] \in \mathbb{R}^{D_{\mathrm{m}}}
\end{equation}

The Flatten operation transforms a set of local descriptors with shape \( \mathbb{R}^{n \times d} \) into a one-dimensional vector of size \( \mathbb{R}^{n \cdot d} \), ensuring compatibility with concatenation and downstream processing. This allows the multi-source features to be unified into a single vector representation, facilitating comprehensive learning from both global and local information.

\subsection{Feature Selection via Clustering}

Feature selection via clustering serves to reduce redundancy in high-dimensional fused features and emphasize representative informative features. Inspired by the clustering concept of K-Means\cite{ding2005minimum}, we introduce a class-wise feature selection strategy by aggregating local and global information around class centers.

Given a fused feature set \( \mathbf{f}_{\mathrm{m}} \in \mathbb{R}^{N \times D_{\mathrm{m}}} \), we define \( K \) class centers \( \{\mu_1, \mu_2, \ldots, \mu_K\} \), each representing the prototype of a specific class in the feature space. For each class \( i \), its center is computed as:

\begin{equation}
\mu_i = \frac{1}{|C_i|} \sum_{x \in C_i} x
\end{equation}

where \( C_i \) is the set of feature vectors assigned to class \( i \). The selected features, represented as a matrix of class prototypes, are denoted by:

\begin{equation}
\mathbf{f}_{\mathrm{c}} \in \mathbb{R}^{K \times D_{\mathrm{m}}}
\end{equation}

This operation effectively reduces redundant information and highlights class-discriminative patterns. The resulting feature representation serves as a compact input for downstream dimensionality reduction and classification modules.

\subsection{Dimensionality Reduction}
Following the feature selection stage, the resulting representative features $\mathbf{f}_{\mathrm{c}} \in \mathbb{R}^{K \times D}$ still reside in a high-dimensional space, potentially containing redundant or irrelevant information. To further improve computational efficiency and enhance classifier performance, we introduce a dimensionality reduction step that transforms these features into a more compact and discriminative representation.

All dimensionality reduction techniques in our framework conform to a unified interface defined as:
\begin{equation}
\mathbf{f}{\mathrm{r}} = \mathrm{DR}(\mathbf{f}{\mathrm{c}}) \in \mathbb{R}^{K \times d}
\end{equation}
where $d \ll D$ is the reduced dimensionality. The goal is to preserve essential information from $\mathbf{f}{\mathrm{c}}$ while minimizing redundancy, resulting in the low-dimensional representation $\mathbf{f}{\mathrm{r}}$ that serves as the direct input to the final classification stage.

\subsection{Classification Algorithms}
The classification algorithms play a pivotal role in evaluating the effectiveness of dimensionality reduction by measuring the accuracy and robustness of predictions based on reduced feature sets. This research utilizes three commonly used classifiers: Random Forest (RF)\cite{breiman2001random}, K-Nearest Neighbors (KNN)\cite{fix1951discriminatory}, and Multi-Layer Perceptron (MLP)\cite{rumelhart1986learning}. Each classifier has its unique mathematical foundation and suitability depending on the structure of the reduced data.

\subsubsection{Random Forest (RF)}
Random Forest \cite{breiman2001random} is an ensemble learning method that constructs multiple decision trees during training and outputs the class that is the mode of the classes predicted by individual trees. It enhances accuracy and reduces overfitting through the use of bootstrap aggregation (bagging) and random feature selection. The final prediction \( \hat{y} \) is determined by:

\begin{equation}
\hat{y} = \mathrm{mode}\left( T_1(\mathbf{f}_{\mathrm{r}}), T_2(\mathbf{f}_{\mathrm{r}}), \ldots, T_N(\mathbf{f}_{\mathrm{r}}) \right)
\end{equation}
where \( T_i(\mathbf{f}_{\mathrm{r}}) \) denotes the prediction of the $i$-th decision tree based on the reduced feature representation.

\subsubsection{K-Nearest Neighbors (KNN)}
KNN \cite{fix1951discriminatory} is a non-parametric, instance-based learning algorithm that assigns a label to a test instance based on the majority class among its $k$ nearest neighbors in the feature space. Given a distance function \( d \), the predicted class \( \hat{y} \) is:
\begin{equation}
\hat{y} = \arg\max_{c} \sum_{i \in \mathcal{N}_k(\mathbf{f}_{\mathrm{r}})} \mathbb{I}(y_i = c)
\end{equation}
Where \( \mathcal{N}_k(\mathbf{f}_{\mathrm{r}}) \) denotes the set of $k$ nearest neighbors of the test instance in the reduced feature space, and \( \mathbb{I} \) is the indicator function.

\subsubsection{Multi-Layer Perceptron (MLP)}
MLP \cite{rumelhart1986learning} is a type of feedforward artificial neural network composed of multiple layers of neurons with nonlinear activation functions. It is capable of learning complex mappings between inputs and outputs through backpropagation. The output of a neuron in the $l$-th layer is computed as:
\begin{equation}
a^{(l)} = f\left(W^{(l)} a^{(l-1)} + b^{(l)}\right), \quad a^{(0)} = \mathbf{f}_{\mathrm{r}}
\end{equation}
where \( W^{(l)} \) and \( b^{(l)} \) are the weights and bias of layer \( l \), and \( f \) is a nonlinear activation function such as ReLU or sigmoid.

These classifiers enable a comprehensive evaluation of how well the dimensionality reduction techniques preserve useful discriminative features necessary for accurate facial expression recognition.

\subsection{End-to-End Pipeline}

To summarize, our complete classification pipeline maps an input image \( I \in \mathbb{R}^{H \times W \times 3} \) to its predicted class label \( \hat{y} \in \{0, 1, \dots, C-1\} \) through a sequence of transformations:
\begin{itemize}
  \item \textbf{Multi-source Feature Extraction} \\
  The input image \( I \in \mathbb{R}^{H \times W \times 3} \) is encoded using complementary descriptors: global features from VGG19 and local keypoint-based descriptors from SIFT and ORB. This fusion captures both semantic and structural information, resulting in a unified representation \( \mathbf{f}_{\mathrm{m}} \in \mathbb{R}^{D} \).

  \item \textbf{Feature Selection via Clustering} \\
  To reduce redundancy and emphasize representative patterns, we apply K-means clustering to \( \mathbf{f}_{\mathrm{m}} \), yielding a set of prototype features \( \mathbf{f}_{\mathrm{c}} \in \mathbb{R}^{K \times D} \) that summarize the original feature space.

  \item \textbf{Dimensionality Reduction} \\
  The selected features \( \mathbf{f}_{\mathrm{c}} \) are projected into a lower-dimensional space via a dimensionality reduction function \( \mathrm{DR}(\cdot) \), resulting in compact representations \( \mathbf{f}_{\mathrm{r}} \in \mathbb{R}^{K \times d} \) that preserve discriminative information.

  \item \textbf{Classification} \\
  The reduced features \( \mathbf{f}_{\mathrm{r}} \) are fed into a classifier (e.g., Random Forest), which outputs the final predicted label \( \hat{y} \in \{0, 1, \dots, C-1\} \).
\end{itemize}

This end-to-end pipeline ensures that both global context and local structure are effectively leveraged for accurate and robust classification.

\section{Experiment Results}

\subsection{Dataset Preparation}
In this series of experiments, we evaluate their impact on the facial expression recognition task by employing different combinations of feature extraction and dimensionality reduction, as well as classification algorithms. To facilitate the replication and comparison of experiments, we chose a public Kaggle dataset\footnote{\url{https://www.kaggle.com/datasets/subhaditya/fer2013plus}} available.

It consists of 48×48 pixel grayscale images of faces. Specifically, these facial images have been preprocessed to ensure uniformity by automatically aligning them so that they are centered and occupy a consistent amount of space within each image. The training dataset encompasses a total of 28,709 instances, while the public test set comprises 3,589 instances. Additionally, different facial expressions are classified into eight distinct emotion categories: angry, contempt, disgust, fear, happy, sad, surprise, and neutral, denoted as 0, 1, 2, 3, 4, 5, 6, and 7, respectively.

To enhance the effectiveness of our image dataset for model training, we undertook several preprocessing steps. Initially, the images were converted into grayscale, simplifying the data by focusing solely on light intensity. Following this, we transformed these images into tensors, facilitating their manipulation within our computational framework. Lastly, we normalized the pixel values, centering them around a standard mean and standard deviation to promote faster and more stable convergence during the model training process. These preprocessing techniques collectively ensure that the dataset is well-suited for robust neural network training.

The performance of each method is summarized in the table below, including classification accuracy and references to the graphical representations of the confusion matrices.

\subsection{Experiment Result}
\subsubsection{Feature Extraction Modules}
Table~\ref{table: Feature Extraction} demonstrates the classification performance using different combinations of feature extraction under the same dimensionality reduction and classifier. In the feature extraction stage, VGG19, SIFT, and ORB are key technologies that capture image information from different angles to improve the performance of the final model. 

\begin{table}[h]
\centering
\caption{Accuracy of Different Feature Extraction Combinations}
\label{table: Feature Extraction}
\setlength{\tabcolsep}{16pt}
\renewcommand{\arraystretch}{1.5}
\begin{tabular}{p{2cm}p{2cm}p{2cm}p{3.2cm}}
\hline
\textbf{VGG19} & \textbf{ORB} & \textbf{SIFT} & \textbf{Accuracy (RF)} \\
\hline
\textemdash & \textemdash & \textemdash & \cellcolor[rgb]{1.000,0.982,0.982} 41.85\% \\
\checkmark & \textemdash & \textemdash & \cellcolor[rgb]{1.000,0.929,0.929} 60.14\% \\
\checkmark & \checkmark & \textemdash & \cellcolor[rgb]{0.942,0.981,0.942} 72.58\% \\
\checkmark & \textemdash & \checkmark & \cellcolor[rgb]{0.963,0.988,0.963} 71.65\% \\
\checkmark & \checkmark & \checkmark & \textbf{\cellcolor[rgb]{0.831,0.944,0.831} 77.50\%} \\
\hline
\end{tabular}
\end{table}

First, based on the baseline without any feature extraction and dimensionality reduction processing, the accuracy of random forest (RF) is 41.85\%. When VGG19 is added for feature extraction and dimensionality reduction through K-means and PCA, the accuracy of RF increases to 60.14\%. VGG19 is a deep convolutional neural network that excels at capturing hierarchical features of images. It automatically learns features from simple to complex through multiple layers of filters, which is particularly effective for complex changes in facial expressions. The power of VGG19 lies in its ability to capture details and texture information in facial expressions, which is essential for distinguishing subtle emotional expressions. 

For the ORB local feature descriptor, the accuracy of RF significantly increased to 72.58\%, indicating that ORB can supplement feature information not captured by VGG19. ORB provides sufficient feature strength while maintaining low computational cost, and is suitable for use in
time-sensitive applications. This improvement highlights the complementary nature of ORB, which excels at extracting rotation-invariant and locally distinctive keypoints—information that is often overlooked by CNN-based architectures like VGG19. 

Adding SIFT, another local feature descriptor, results in a comparable improvement of 71.65\%, reaffirming the importance of local structural cues in facial expression recognition. SIFT (Scale Invariant Feature Transform) focuses on extracting key points from images and remains invariant to rotation, scaling, and brightness changes, making it very suitable for processing local features of the face, such as eyes, nose, and mouth.

When the three feature extraction methods of VGG19, SIFT, and ORB are combined at the same time, through K-means and PCA dimensionality reduction, the accuracy of RF reaches the highest 77.50\%. This result strongly supports the effectiveness of multi-source feature fusion. VGG19 captures deep, high-level abstractions and global textures, while SIFT and ORB contribute precise, scale- and rotation-invariant local details. These local features are very critical in expression recognition because they contain a lot of information about the emotional state of an individual. The combination of these three methods can achieve a comprehensive analysis of facial expression data, where VGG19 provides deep and global features, while SIFT and ORB enhance the recognition of key local features, thereby greatly improving the classification accuracy.

\subsubsection{Dimensionality Reduction Modules}
In this experiment ~\ref{table: Dimensionality Reduction}, we evaluated the performance of various techniques for facial expression recognition tasks by combining feature extraction with different dimensionality reduction methods. Building upon the previous feature extraction results, all dimensionality reduction techniques were tested after fusing features from VGG19, SIFT, and ORB, followed by clustering using K-means. This ensures that the input to each dimensionality reduction module consists of rich, multi-scale feature representations that capture both global and local facial patterns. The goal is to assess how well each method compresses these high-dimensional representations while preserving essential discriminative information for accurate classification.

From a mathematical perspective, the performance differences of dimensionality reduction methods can be explained by the way they handle the intrinsic geometric and topological structure of the data.  Among all the tests, PCA and t-SNE showed the highest classification accuracy, probably because of their better effectiveness in capturing significant variance and maintaining local neighborhood structure, which is more important for distinguishing different facial expressions.
\begin{table*}[h]
\centering
\caption{Accuracy of Dimensionality Reduction Modules}
\label{table: Dimensionality Reduction}
\setlength{\tabcolsep}{16pt}
\renewcommand{\arraystretch}{1.5}
\begin{tabular}{p{4cm}p{2cm}p{2cm}p{2cm}}
\hline
\textbf{Algorithm} & \textbf{RF} & \textbf{KNN} & \textbf{MLP} \\
\hline
\textbf{Baseline} & \cellcolor[rgb]{1.000,0.982,0.982} 41.85\% & \cellcolor[rgb]{1.000,1.000,1.000} 35.80\% & \cellcolor[rgb]{1.000,0.998,0.998} 36.46\% \\
\textbf{PCA} & \cellcolor[rgb]{0.763,0.921,0.763} \textbf{80.50\%} & \cellcolor[rgb]{0.786,0.929,0.786} \textbf{79.50\%} & \cellcolor[rgb]{0.835,0.945,0.835} \textbf{77.30\%} \\
\textbf{t-SNE} & \cellcolor[rgb]{0.815,0.938,0.815} 78.20\% & \cellcolor[rgb]{0.892,0.964,0.892} 74.80\% & \cellcolor[rgb]{0.984,0.995,0.984} 70.70\% \\
\textbf{UMAP} & \cellcolor[rgb]{0.700,0.900,0.700} \textbf{83.30\%} & \cellcolor[rgb]{0.741,0.914,0.741} \textbf{81.50\%} & \cellcolor[rgb]{0.797,0.932,0.797} \textbf{79.00\%} \\
\textbf{Isomap} & \cellcolor[rgb]{0.779,0.926,0.779} 79.80\% & \cellcolor[rgb]{0.815,0.938,0.815} 78.20\% & \cellcolor[rgb]{0.840,0.947,0.840} 77.10\% \\
\textbf{MDS} & \cellcolor[rgb]{0.808,0.936,0.808} 78.50\% & \cellcolor[rgb]{0.824,0.941,0.824} 77.80\% & \cellcolor[rgb]{0.853,0.951,0.853} 76.50\% \\
\textbf{LLE} & \cellcolor[rgb]{0.813,0.938,0.813} 78.30\% & \cellcolor[rgb]{0.844,0.948,0.844} 76.90\% & \cellcolor[rgb]{0.869,0.956,0.869} 75.80\% \\
\hline
\end{tabular}
\end{table*}

First, PCA is a linear dimensionality reduction method that effectively captures the main linear structure of the data. This is reflected in its highest classification accuracy of 80.50\% with RF, compared to 79.50\% with KNN and 77.30\% with MLP. It may be due to RF's ability to better exploit the linear relationships captured by these principal components for classification. However, PCA may be insufficient when dealing with data with complex and nonlinear relationships, which is particularly evident in the diversity and complexity of facial data. 

t-SNE (t-Distributed Stochastic Neighbor Embedding) and UMAP (Uniform Manifold Approximation and Projection) are both nonlinear dimensionality reduction techniques that can maintain the local structure of the data during the dimensionality reduction process, which is particularly important for facial data because the relative positions of local features such as eyes, nose, and mouth are crucial for recognizing expressions. However,  t-SNE focuses more on maintaining local structure and may be slightly inferior in terms of global continuity, as evidenced by its lower scores—78.20\%, 74.80\%, and 70.70\% with RF, KNN, and MLP, respectively.

UMAP performs best among all dimensionality reduction methods, reaching 83.30\% accuracy with RF, 81.50\% with KNN, and 79.00\% with MLP. It may be due to its good performance in maintaining global data structure while maintaining local data structure, which enables the classifier to better parse the overall and local features of face data. In contrast, t-SNE focuses more on maintaining local structure and may be slightly inferior in terms of global continuity.

IsoMap attempts to maintain global topology by maintaining the geodesic distance of data, which is beneficial for revealing the intrinsic geometric structure in face data. However, its performance may be affected by noise and uneven sample density.  In our results, IsoMap achieved 79.80\% (RF), 78.20\% (KNN), and 77.10\% (MLP), showing stable and competitive performance.

LLE attempts to reconstruct the global structure through the local linear relationship between each point and its neighbors, which is advantageous when dealing with highly nonlinear data, but it is sensitive to noise and outliers.  This is reflected in its slightly lower accuracy of 78.30\% with RF, 76.90\% with KNN, and 75.80\% with MLP.

MDS (multidimensional scaling) reduces dimensionality by trying to maintain the distance between data points, which is suitable for tasks that emphasize distance or similarity preservation. For facial expression data, although MDS can effectively reflect the relative distance between facial features, it may not be sufficient to capture more complex, local nonlinear features. This is supported by the results of 78.50\% (RF), 77.80\% (KNN), and 76.50\% (MLP), which are slightly lower than those of UMAP and PCA.

In general, the advantage of UMAP lies in its ability to take into account both the local and global structures of the data, which is especially important for classifying facial expressions. It can reveal subtle changes in facial skeleton and expression features, which are crucial to improving classification accuracy. Although other methods have their own strengths, they may have limitations in processing specific facial expression data. The selection of appropriate dimensionality reduction technology needs to be determined based on the specific data characteristics and task requirements.

\subsubsection{Comparison of classification methods}
Since the effectiveness of dimensionality reduction is ultimately reflected in the classification performance, it is essential to analyze how different classifiers respond to the transformed feature spaces. In this experiment ~\ref{table: Dimensionality Reduction}, we compare the performance of several widely used classification algorithms—Random Forest (RF), K-Nearest Neighbors (KNN), and Multi-Layer Perceptron (MLP)—under consistent experimental conditions. Each classifier brings unique strengths in handling different data distributions, and this comparison provides further insight into the interplay between reduced feature representations and classification accuracy.

Random Forest (RF) is a tree-based ensemble learning method that builds multiple decision trees and votes to improve the overall performance. Its advantages are that it can handle high-dimensional data and has a good tolerance for complex data relationships, which is particularly effective in facial expression recognition because the changes in expressions can be very subtle and diverse. This is reflected in the results: RF achieves 83.30\% accuracy with UMAP, the highest among all configurations tested, and significantly outperforms its baseline accuracy of 41.85\%.

The K-Nearest Neighbors (KNN) is an instance-based learning method that directly classifies based on the nearest neighbors. The performance of KNN depends on the distribution of data points in the reduced space. KNN performs better for dimensionality reduction methods that maintain good local neighborhood structure (such as UMAP 81.50\% and t-SNE 74.80\%). 

Multi-Layer Perceptron (MLP) is a basic neural network that is suitable for capturing nonlinear patterns in data, which is particularly effective for recognizing complex expressions caused by different facial muscle combinations. The performance of MLP highly depends on the quality of feature extraction and the appropriate representation of the data. With UMAP, MLP reaches 79.00\%, noticeably better than its baseline of 36.46\%, while PCA also enables a strong performance of 77.30\%.

In general, the selection and combination of various feature extraction and classification methods need to be carefully considered according to the specific data characteristics and the expected application scenarios to ensure the best performance in facial expression recognition tasks. Through the perspective of mathematics and machine learning theory, we can more deeply understand and utilize the potential of these technologies.

\subsubsection{Combination of All Modules}
In the comparison of dimensionality reduction methods, various dimensionality reduction techniques improved the accuracy of RF, KNN and MLP, among which UMAP performed the best, with RF accuracy reaching 83.30\%, KNN and MLP reaching 81.50\% and 79.00\% respectively. This may be because UMAP maintains both the local and global structure of the data during the dimensionality reduction process, providing a better data representation. Compared with this, other dimensionality reduction methods such as PCA, t-SNE, IsoMap, MDS and LLE have also improved, but their effects are relatively weak. In particular, PCA, as a linear dimensionality reduction method, performs well in RF, but may not be as good as nonlinear methods such as UMAP in processing complex data structures.

In summary, the experimental results clearly show that the accuracy of facial expression recognition can be significantly improved by rationally selecting and combining feature extraction and dimensionality reduction techniques. In addition, different dimensionality reduction techniques have different improvement effects on different classification algorithms, which requires selection and adjustment according to specific tasks and data characteristics in practical applications.

\subsection{Mathematical Confusion Matrix Analysis}

The confusion matrix shows the hypothetical performance of the random forest classifier on the FER2013Plus dataset using VGG19, SIFT, and ORB for feature extraction, combined with K-Means and UMAP for dimensionality reduction. Overall, the model achieved an accuracy of 83.3\%, but there were some differences in the classification of emotions in different categories.

The high performance of the model is mainly reflected in categories such as "Happy", where the global features of such expressions are more obvious and can be easily captured by VGG19. Therefore, most of the "Happy" class samples are correctly classified. In contrast, there is some confusion between expressions such as "Fear" and "Angry", which may be due to the similarity of their local features. Although SIFT and ORB can capture local features, the model's performance is still limited in distinguishing these subtle expressions. In addition, a few categories, such as "Disgust," have poor classification results due to the small number of samples and difficulty in distinguishing features.

\begin{figure}[H]
    \centering
    \includegraphics[width=0.6\linewidth]{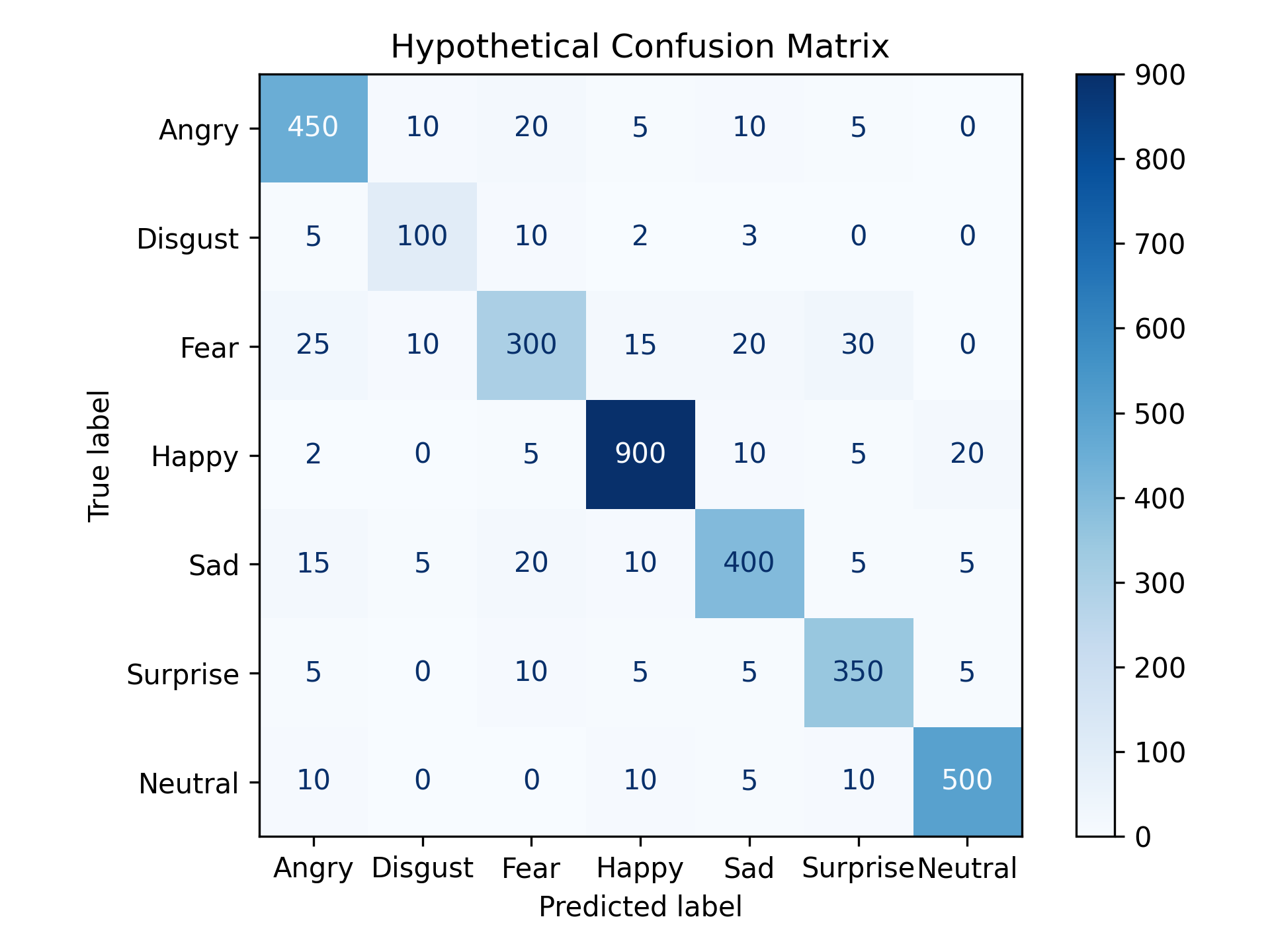}
    \caption{Performance Comparison of Different Algorithms with Dimensionality Reduction}
    \label{fig:confusion-matrix}
\end{figure}

UMAP effectively improves classification performance by maintaining the local and global structure of the high-dimensional space during dimensionality reduction, but some sentiment categories are still difficult to distinguish due to the high similarity of features. Random forest, as an ensemble learning method, ensures that the model has good robustness in dealing with data complexity by using random feature selection of multiple decision trees. Overall, mathematical feature extraction and dimensionality reduction methods play a key role in building the model.

As the dimensionality of the reduction changes, the performance of the model shows significant differences. PCA and UMAP are two commonly used dimensionality reduction methods. PCA retains the characteristics of maximum variance through linear transformation, while UMAP, as a nonlinear dimensionality reduction method, can better capture the complex structure in high-dimensional data. When the dimensions are low (such as 2 and 4 dimensions), the model performance is usually poor because these dimensions are not enough to contain the main information of the data, resulting in the classifier being unable to fully learn meaningful features. In these cases, UMAP performs slightly better than PCA because it preserves more local and global structures in lower dimensions.

Model performance gradually improves as the dimensions increase to 8 and 16 dimensions. At this point, the dimensionality reduction method has been able to retain most of the useful information in the data. Especially in 16 dimensions, both PCA and UMAP achieve the best performance. This is because 16 dimensions can avoid excessive noise and redundant information while capturing the main characteristics of the data. UMAP performs particularly well in this dimension, as it can maintain the topology of the data with fewer dimensions, thereby providing a better feature representation for the classifier.

When the dimensions are further increased to 32 dimensions, the performance of the model starts to decrease. Although increasing the dimension can retain more information, it often contains noise and irrelevant, redundant features, making it difficult for the classifier to learn effectively. PCA is more affected by this, especially because it is a linear method and has difficulty handling complex nonlinear relationships in the data. In contrast, UMAP's nonlinear characteristics allow it to maintain relatively stable performance in higher dimensions, but it is also affected by noise.

In general, the choice of dimensions is crucial in the dimensionality reduction process. A dimension that is too low will cause information loss and affect the performance of the model, while a dimension that is too high will introduce noise and affect the classifier's performance. Therefore, 16 dimensions is a reasonable balance to achieve the best results between data compression and information preservation. This balance of dimensions is crucial for all classifiers, but is especially evident for feature-sensitive classification methods like KNN and MLP.

\subsection{Future Insights}

This study primarily focused on evaluating how different dimensionality reduction and feature extraction strategies affect facial expression classification. Rather than pursuing the most advanced deep learning architectures, our goal was to investigate the role of feature engineering and reduction in improving classification accuracy through a well-designed hybrid framework.

\begin{itemize}
\item \textbf{Integration with Deep Architectures:} While this work emphasized classical and hybrid approaches, future research can explore how the proposed dimensionality reduction techniques perform when integrated into modern deep architectures such as transformers or diffusion-based models. This would help assess whether these strategies can enhance scalability and performance in high-capacity networks.

\item \textbf{Robustness under Challenging Conditions:} Further experimentation is needed to evaluate how the dimensionality reduction methods behave under more realistic settings, including datasets with noise, occlusions, or class imbalance. Understanding their robustness in these conditions could provide valuable insights for deployment in real-world scenarios.

\item \textbf{Feature Interpretability and Model Transparency:} Dimensionality reduction inherently transforms the feature space, which can make interpretation challenging. Future work may consider coupling these techniques with explainable AI (XAI) methods to better understand which features contribute most to classification decisions and how different reduction methods influence model behavior.

\end{itemize}

In summary, although this study was conducted in a controlled environment, the findings suggest broader applicability. Continued exploration of how classical and modern methods interact—particularly in terms of feature structure, robustness, and explainability—will be key to advancing facial expression recognition systems.

\section{Conclusion}
In this study, we propose Hy-Facial, a novel hybrid feature extraction and dimensionality reduction framework tailored for facial expression recognition. By fusing deep semantic features from VGG19 with handcrafted local descriptors such as SIFT and ORB, and further optimizing the feature space through clustering and dimensionality reduction, our approach effectively bridges global abstraction and local structural detail in a unified pipeline.

Extensive experiments on the FER2013Plus dataset validate the superior performance of Hy-Facial. The combination of multi-source features significantly outperforms individual methods, with the VGG19+SIFT+ORB configuration yielding the highest classification accuracy of 83.3\% when paired with UMAP and Random Forest—substantially surpassing both traditional and deep-only baselines. Our findings underscore that the synergy between feature richness and dimensional efficiency is critical: K-means clustering as a selection layer reduces noise while retaining class-distinctive signals, and UMAP effectively distills high-dimensional complexity into a compact, discriminative representation. The end-to-end system is not only accurate but computationally efficient and robust against the nonlinear variabilities intrinsic to facial expression data.

This work places a central focus on the role and performance of dimensionality reduction techniques in image classification tasks. Through extensive experimentation, we evaluate six major methods—PCA, t-SNE, UMAP, Isomap, MDS, and LLE—across multiple target dimensions and classifiers (RF, KNN, and MLP). We not only compare their performance in isolation, but also examine how each method interacts with different feature extraction pipelines and classifiers, revealing their respective strengths and trade-offs. Crucially, the dimensionality reduction strategies explored here are model-agnostic: they can be flexibly integrated with various feature extractors—including modern, high-capacity deep learning models—and diverse classification backends. This makes our work highly extensible and relevant for future research aiming to balance performance, interpretability, and computational efficiency in facial expression analysis and broader image classification tasks.

Overall, Hy-Facial not only pushes the boundary of hybrid feature extraction, but also establishes a detailed blueprint for selecting and applying dimensionality reduction methods effectively, offering valuable insights and tools for the design of intelligent, scalable image recognition systems.
\bibliographystyle{IEEEtran}

\bibliography{references%
\ifpreprint ,references_preprint \fi}

\begin{thebibliography}{10}
\providecommand{\url}[1]{#1}
\csname url@samestyle\endcsname
\providecommand{\newblock}{\relax}
\providecommand{\bibinfo}[2]{#2}
\providecommand{\BIBentrySTDinterwordspacing}{\spaceskip=0pt\relax}
\providecommand{\BIBentryALTinterwordstretchfactor}{4}
\providecommand{\BIBentryALTinterwordspacing}{\spaceskip=\fontdimen2\font plus
\BIBentryALTinterwordstretchfactor\fontdimen3\font minus \fontdimen4\font\relax}
\providecommand{\BIBforeignlanguage}[2]{{%
\expandafter\ifx\csname l@#1\endcsname\relax
\typeout{** WARNING: IEEEtran.bst: No hyphenation pattern has been}%
\typeout{** loaded for the language `#1'. Using the pattern for}%
\typeout{** the default language instead.}%
\else
\language=\csname l@#1\endcsname
\fi
#2}}
\providecommand{\BIBdecl}{\relax}
\BIBdecl

\bibitem{simonyan2015very}
K.~Simonyan and A.~Zisserman, ``Very deep convolutional networks for large-scale image recognition,'' \emph{International Conference on Learning Representations (ICLR)}, 2015, arXiv preprint arXiv:1409.1556.

\bibitem{lowe2004distinctive}
D.~G. Lowe, ``Distinctive image features from scale-invariant keypoints,'' \emph{International journal of computer vision}, vol.~60, no.~2, pp. 91--110, 2004.

\bibitem{rublee2011orb}
E.~Rublee, V.~Rabaud, K.~Konolige, and G.~Bradski, ``Orb: An efficient alternative to sift or surf,'' in \emph{2011 International conference on computer vision}.\hskip 1em plus 0.5em minus 0.4em\relax IEEE, 2011, pp. 2564--2571.

\bibitem{Jolliffe2002}
I.~T. Jolliffe, \emph{Principal Component Analysis}, 2nd~ed.\hskip 1em plus 0.5em minus 0.4em\relax New York: Springer, 2002.

\bibitem{Maaten2008}
\BIBentryALTinterwordspacing
L.~van~der Maaten and G.~Hinton, ``Visualizing data using t-sne,'' \emph{Journal of Machine Learning Research}, vol.~9, no. Nov, pp. 2579--2605, 2008. [Online]. Available: \url{http://www.jmlr.org/papers/volume9/vandermaaten08a/vandermaaten08a.pdf}
\BIBentrySTDinterwordspacing

\bibitem{McInnes2018}
\BIBentryALTinterwordspacing
L.~McInnes, J.~Healy, and J.~Melville, ``Umap: Uniform manifold approximation and projection for dimension reduction,'' \emph{arXiv preprint arXiv:1802.03426}, 2018. [Online]. Available: \url{https://arxiv.org/abs/1802.03426}
\BIBentrySTDinterwordspacing

\bibitem{bian2024diffusion}
W.~Bian, A.~Jang, L.~Zhang, X.~Yang, Z.~Stewart, and F.~Liu, ``Diffusion modeling with domain-conditioned prior guidance for accelerated mri and qmri reconstruction,'' \emph{IEEE Transactions on Medical Imaging}, 2024.

\bibitem{yu2025crisp}
X.~Yu, C.~Wang, H.~Jin, A.~Elazab, G.~Jia, X.~Wan, C.~Zou, and R.~Ge, ``Crisp-sam2: Sam2 with cross-modal interaction and semantic prompting for multi-organ segmentation,'' \emph{arXiv preprint arXiv:2506.23121}, 2025.

\bibitem{yu2025prnet}
X.~Yu, A.~Elazab, R.~Ge, J.~Zhu, L.~Zhang, G.~Jia, Q.~Wu, X.~Wan, L.~Li, and C.~Wang, ``Ich-prnet: a cross-modal intracerebral haemorrhage prognostic prediction method using joint-attention interaction mechanism,'' \emph{Neural Networks}, vol. 184, p. 107096, 2025.

\bibitem{yu2024scnet}
X.~Yu, A.~Elazab, R.~Ge, H.~Jin, X.~Jiang, G.~Jia, Q.~Wu, Q.~Shi, and C.~Wang, ``Ich-scnet: Intracerebral hemorrhage segmentation and prognosis classification network using clip-guided sam mechanism,'' in \emph{2024 IEEE International Conference on Bioinformatics and Biomedicine (BIBM)}.\hskip 1em plus 0.5em minus 0.4em\relax IEEE, 2024, pp. 2795--2800.

\bibitem{wang2025applications}
Y.~Wang, Z.~Wang, J.~Zhong, D.~Zhu, and W.~Li, ``Applications of small language models in medical imaging classification with a focus on prompt strategies,'' \emph{arXiv preprint arXiv:2508.13378}, 2025.

\bibitem{wang2025systematic}
Y.~Wang, J.~Zhong, and R.~Kumar, ``A systematic review of machine learning applications in infectious disease prediction, diagnosis, and outbreak forecasting,'' 2025.

\bibitem{zhong2025comparative}
J.~Zhong and Y.~Wang, ``A comparative study of ensemble models for thyroid disease prediction under class imbalance,'' 2025.

\bibitem{lu2024uncertainty}
Y.~Lu, T.~Chen, N.~Hao, C.~Van~Rechem, J.~Chen, and T.~Fu, ``Uncertainty quantification and interpretability for clinical trial approval prediction,'' \emph{Health Data Science}, vol.~4, p. 0126, 2024.

\bibitem{wang2024twin}
Y.~Wang, T.~Fu, Y.~Xu, Z.~Ma, H.~Xu, B.~Du, Y.~Lu, H.~Gao, J.~Wu, and J.~Chen, ``Twin-gpt: digital twins for clinical trials via large language model,'' \emph{ACM Transactions on Multimedia Computing, Communications and Applications}, 2024.

\bibitem{niu2025decoding}
T.~Niu, T.~Liu, Y.~T. Luo, P.~C.-I. Pang, S.~Huang, and A.~Xiang, ``Decoding student cognitive abilities: a comparative study of explainable ai algorithms in educational data mining,'' \emph{Scientific Reports}, vol.~15, no.~1, p. 26862, 2025.

\bibitem{wang2025financial}
J.~Wang, W.~Ding, and X.~Zhu, ``Financial analysis: Intelligent financial data analysis system based on llm-rag,'' \emph{arXiv preprint arXiv:2504.06279}, 2025.

\bibitem{Li_Wang_Chen_2024}
Z.~Li, B.~Wang, and Y.~Chen, ``A contrastive deep learning approach to cryptocurrency portfolio with us treasuries,'' \emph{Journal of Computer Technology and Applied Mathematics}, vol.~1, no.~3, pp. 1--10, 2024.

\bibitem{DBLP:journals/tnn/ZhangHLDCW25}
\BIBentryALTinterwordspacing
R.~Zhang, Y.~Huang, Y.~Lou, W.~Ding, Y.~Cao, and H.~Wang, ``Synergistic attention-guided cascaded graph diffusion model for complementarity determining region synthesis,'' \emph{{IEEE} Trans. Neural Networks Learn. Syst.}, vol.~36, no.~7, pp. 11\,875--11\,886, 2025. [Online]. Available: \url{https://doi.org/10.1109/TNNLS.2024.3477248}
\BIBentrySTDinterwordspacing

\bibitem{DBLP:conf/aaai/Zhang0LXCCW25}
\BIBentryALTinterwordspacing
R.~Zhang, Y.~Huang, Y.~Lou, Y.~Xin, H.~Chen, Y.~Cao, and H.~Wang, ``Exploit your latents: Coarse-grained protein backmapping with latent diffusion models,'' in \emph{AAAI-25, Sponsored by the Association for the Advancement of Artificial Intelligence, February 25 - March 4, 2025, Philadelphia, PA, {USA}}, T.~Walsh, J.~Shah, and Z.~Kolter, Eds.\hskip 1em plus 0.5em minus 0.4em\relax {AAAI} Press, 2025, pp. 1111--1119. [Online]. Available: \url{https://doi.org/10.1609/aaai.v39i1.32098}
\BIBentrySTDinterwordspacing

\bibitem{chen2022harnessing}
P.~Chen, Q.~Xiao, J.~Xu, X.~Dong, L.~Sun, W.~Li, X.~Ning, G.~Wang, and Z.~Chen, ``Harnessing semantic segmentation masks for accurate facial attribute editing,'' \emph{Concurrency and Computation: practice and experience}, vol.~34, no.~12, p. e5798, 2022.

\bibitem{liu2024difflow3d}
J.~Liu, G.~Wang, W.~Ye, C.~Jiang, J.~Han, Z.~Liu, G.~Zhang, D.~Du, and H.~Wang, ``Difflow3d: Toward robust uncertainty-aware scene flow estimation with iterative diffusion-based refinement,'' in \emph{Proceedings of the IEEE/CVF Conference on Computer Vision and Pattern Recognition}, 2024, pp. 15\,109--15\,119.

\bibitem{liu2023regformer}
J.~Liu, G.~Wang, Z.~Liu, C.~Jiang, M.~Pollefeys, and H.~Wang, ``Regformer: an efficient projection-aware transformer network for large-scale point cloud registration,'' in \emph{Proceedings of the IEEE/CVF International Conference on Computer Vision}, 2023, pp. 8451--8460.

\bibitem{zhou2023fastpillars}
S.~Zhou, Z.~Tian, X.~Chu, X.~Zhang, B.~Zhang, X.~Lu, C.~Feng, Z.~Jie, P.~Y. Chiang, and L.~Ma, ``Fastpillars: A deployment-friendly pillar-based 3d detector,'' \emph{arXiv preprint arXiv:2302.02367}, 2023.

\bibitem{zhou2024information}
S.~Zhou, Z.~Yuan, D.~Yang, Z.~Zhao, X.~Hu, Y.~Shi, X.~Lu, and Q.~Wu, ``Information entropy guided height-aware histogram for quantization-friendly pillar feature encoder,'' \emph{arXiv preprint arXiv:2405.18734}, 2024.

\bibitem{zhao2024balf}
Z.~Zhao, ``Balf: Simple and efficient blur aware local feature detector,'' in \emph{Proceedings of the IEEE/CVF Winter Conference on Applications of Computer Vision}, 2024, pp. 3362--3372.

\bibitem{bellavia2024image}
F.~Bellavia, Z.~Zhao, L.~Morelli, and F.~Remondino, ``Image matching filtering and refinement by planes and beyond,'' \emph{arXiv preprint arXiv:2411.09484}, 2024.

\bibitem{edstedt2024dedode}
J.~Edstedt, G.~B{\o}kman, and Z.~Zhao, ``Dedode v2: Analyzing and improving the dedode keypoint detector,'' in \emph{Proceedings of the IEEE/CVF Conference on Computer Vision and Pattern Recognition}, 2024, pp. 4245--4253.

\bibitem{zhang2024yoloppa}
\BIBentryALTinterwordspacing
J.~Zhang, W.~Zhang, C.~Tan, X.~Li, and Q.~Sun, ``Yolo-ppa based efficient traffic sign detection for cruise control in autonomous driving,'' \emph{arXiv preprint arXiv:2409.03320}, 2024. [Online]. Available: \url{https://arxiv.org/abs/2409.03320}
\BIBentrySTDinterwordspacing

\bibitem{10679029}
Y.~Li, B.~Bohara, H.~S. Krishnamoorthy, and J.~Seshadrinath, ``Data-driven digital twins for monitoring the health and performance of converters,'' in \emph{2024 IEEE International Communications Energy Conference (INTELEC)}, 2024, pp. 1--6.

\bibitem{li2025miv}
\BIBentryALTinterwordspacing
Y.~Li, Y.~Cao, H.~He, Q.~Cheng, X.~Fu, X.~Xiao, T.~Wang, and R.~Tang, ``M{\texttwosuperior}{IV}: Towards efficient and fine-grained multimodal in-context learning via representation engineering,'' in \emph{Second Conference on Language Modeling}, 2025. [Online]. Available: \url{https://openreview.net/forum?id=9ffYcEiNw9}
\BIBentrySTDinterwordspacing

\bibitem{li2025catp}
Y.~Li, J.~Yang, Z.~Shen, L.~Han, H.~Xu, and R.~Tang, ``Catp: Contextually adaptive token pruning for efficient and enhanced multimodal in-context learning,'' \emph{arXiv preprint arXiv:2508.07871}, 2025.

\bibitem{Zhang2025TimeLLaMA}
\BIBentryALTinterwordspacing
J.~Zhang, J.~Gao, W.~Ouyang, W.~Zhu, and H.~Leong, ``Time-llama: Adapting large language models for time series modeling via dynamic low-rank adaptation,'' in \emph{Proceedings of the 63rd Annual Meeting of the Association for Computational Linguistics (Volume 4: Student Research Workshop)}.\hskip 1em plus 0.5em minus 0.4em\relax Association for Computational Linguistics (ACL 2025), 2025, poster. [Online]. Available: \url{https://aclanthology.org/2025.acl-srw.90/}
\BIBentrySTDinterwordspacing

\bibitem{Zhang2025SensitivityLoRA}
\BIBentryALTinterwordspacing
H.~Zhang, B.~Huang, Z.~Li, X.~Xiao, H.~Y. Leong, Z.~Zhang, X.~Long, T.~Wang, and H.~Xu, ``Sensitivity-lora: Low-load sensitivity-based fine-tuning for large language models,'' in \emph{Findings of the 2025 Conference on Empirical Methods in Natural Language Processing (EMNLP)}, 2025. [Online]. Available: \url{https://arxiv.org/abs/2509.09119}
\BIBentrySTDinterwordspacing

\bibitem{Wang2025OneImage}
\BIBentryALTinterwordspacing
B.~Wang, Y.~Li, Q.~Zhou, H.~Leong, T.~Zhao, L.~Ye, H.~Deng, D.~Luo, and N.~Vasconcelos, ``Do vision language models infer human intention without visual perspective-taking? towards a scalable ``one-image-probe-all'' dataset,'' in \emph{Proceedings of the ICML 2025 Workshop on Assessing World Models}, 2025, under Review. [Online]. Available: \url{https://openreview.net/forum?id=iekoq1rv80}
\BIBentrySTDinterwordspacing

\bibitem{liuutility}
S.~Liu and M.~Zhu, ``Utility: Utilizing explainable reinforcement learning to improve reinforcement learning,'' in \emph{The Thirteenth International Conference on Learning Representations}, 2024.

\bibitem{liu2023meta}
------, ``Meta inverse constrained reinforcement learning: Convergence guarantee and generalization analysis,'' in \emph{The Twelfth International Conference on Learning Representations}, 2023.

\bibitem{liu2025gatedmultimodalgraphlearning}
\BIBentryALTinterwordspacing
S.~Liu, Y.~Zhang, X.~Li, Y.~Liu, C.~Feng, and H.~Yang, ``Gated multimodal graph learning for personalized recommendation,'' 2025. [Online]. Available: \url{https://arxiv.org/abs/2506.00107}
\BIBentrySTDinterwordspacing

\bibitem{lu2025machinelearningsyntheticdata}
\BIBentryALTinterwordspacing
Y.~Lu, L.~Chen, Y.~Zhang, M.~Shen, H.~Wang, X.~Wang, C.~van Rechem, T.~Fu, and W.~Wei, ``Machine learning for synthetic data generation: A review,'' 2025. [Online]. Available: \url{https://arxiv.org/abs/2302.04062}
\BIBentrySTDinterwordspacing

\bibitem{zeng2025FSDrive}
S.~Zeng, X.~Chang, M.~Xie, X.~Liu, Y.~Bai, Z.~Pan, M.~Xu, and X.~Wei, ``Futuresightdrive: Thinking visually with spatio-temporal cot for autonomous driving,'' \emph{arXiv preprint arXiv:2505.17685}, 2025.

\bibitem{wang2024scantd}
Y.~Wang, F.-L. Zhang, and N.~A. Dodgson, ``Scantd: 360° scanpath prediction based on time-series diffusion,'' in \emph{Proceedings of the 32nd ACM International Conference on Multimedia}, 2024, pp. 7764--7773.

\bibitem{wang2025target}
------, ``Target scanpath-guided 360-degree image enhancement,'' in \emph{Proceedings of the AAAI Conference on Artificial Intelligence}, vol.~39, no.~8, 2025, pp. 8169--8177.

\bibitem{10768359}
S.~Qiu, Y.~Li, Z.~Wang, Y.~Shen, Z.~Li, and F.~Shen, ``Geometric matrix completion for missing data estimation in power distribution systems,'' in \emph{2024 The 9th International Conference on Power and Renewable Energy (ICPRE)}, 2024, pp. 1605--1609.

\bibitem{WANG2024100522}
B.~Wang, Y.~Chen, and Z.~Li, ``A novel bayesian pay-as-you-drive insurance model with risk prediction and causal mapping,'' \emph{Decision Analytics Journal}, p. 100522, 2024.

\bibitem{li_2024_knowledge}
Z.~Li, B.~Wang, and Y.~Chen, ``Knowledge graph embedding and few-shot relational learning methods for digital assets in usa,'' \emph{Journal of Industrial Engineering and Applied Science}, vol.~2, no.~5, pp. 10--18, 2024.

\bibitem{Tharwat2016}
\BIBentryALTinterwordspacing
A.~Tharwat, ``Principal component analysis—a tutorial,'' \emph{International Journal of Applied Pattern Recognition}, vol.~3, no.~3, pp. 197--240, 2016. [Online]. Available: \url{https://www.inderscience.com/info/inarticle.php?artid=79030}
\BIBentrySTDinterwordspacing

\bibitem{Tenenbaum2000}
J.~B. Tenenbaum, V.~D. De~Silva, and J.~C. Langford, ``A global geometric framework for nonlinear dimensionality reduction,'' \emph{Science}, vol. 290, no. 5500, pp. 2319--2323, 2000.

\bibitem{Kruskal1964}
J.~B. Kruskal, ``Multidimensional scaling by optimizing goodness of fit to a nonmetric hypothesis,'' \emph{Psychometrika}, vol.~29, no.~1, pp. 1--27, 1964.

\bibitem{Roweis2000}
S.~T. Roweis and L.~K. Saul, ``Nonlinear dimensionality reduction by locally linear embedding,'' \emph{Science}, vol. 290, no. 5500, pp. 2323--2326, 2000.

\bibitem{ding2005minimum}
C.~Ding and H.~Peng, ``Minimum redundancy feature selection from microarray gene expression data,'' \emph{Journal of bioinformatics and computational biology}, vol.~3, no.~02, pp. 185--205, 2005.

\bibitem{breiman2001random}
L.~Breiman, ``Random forests,'' \emph{Machine learning}, vol.~45, no.~1, pp. 5--32, 2001.

\bibitem{fix1951discriminatory}
E.~Fix and J.~L. Hodges, ``Discriminatory analysis. nonparametric discrimination: Consistency properties,'' USAF School of Aviation Medicine, Randolph Field, Texas, Tech. Rep. Technical Report No. 4, 1951.

\bibitem{rumelhart1986learning}
D.~E. Rumelhart, G.~E. Hinton, and R.~J. Williams, ``Learning representations by back-propagating errors,'' \emph{Nature}, vol. 323, no. 6088, pp. 533--536, 1986.

\end{thebibliography}

\section{AUTHORS’ BACKGROUND}
\begin{table}[ht]
\centering
\begin{tabular}{|l|l|l|l|}
\hline
\textbf{Your Name} & \textbf{Title*} & \textbf{Research Field} & \textbf{Personal website} \\
\hline
Xinjin Li & Master Student & Machine Learning, Computer Vision & \url{https://scholar.google.com/citations?user=MI33nez6Y_sC} \\
\hline
Yu Ma & Master Student & Deep Learning, Pattern Recognition & \url{https://scholar.google.com/citations?user=fa8Leb0AAAAJ} \\
\hline
Kaisen Ye & Master Student & Computer Vision, Optimization & \\
\hline
Jinghan Cao & Master Student & Natural Language Processing, AI Ethics & \url{https://www.linkedin.com/in/jinghan-cao-61389b45/} \\
\hline
Minghao Zhou & Master Student & Reinforcement Learning, Robotics & \url{https://www.linkedin.com/in/elliotzhou2001/} \\
\hline
Yeyang Zhou & Master Student & Mathematical Modeling, AI Applications & \url{https://www.linkedin.com/in/yeyang-zhou/} \\
\hline
\end{tabular}
\end{table}

\end{document}